\documentclass{article}

\usepackage{microtype}
\usepackage{amsmath}
\usepackage{tikz}
\usepackage{mathpazo}
\usetikzlibrary{patterns}
\usepackage[square,numbers]{natbib}

\usepackage{url}
\usepackage{booktabs}
\usepackage{graphicx}
\usepackage{subfigure}
\usepackage{booktabs} 

\usepackage{hyperref}

\usepackage[accepted]{icml2019}

\icmltitlerunning{A General Framework for Fair Regression}

\begin{document}

\twocolumn[
\icmltitle{A General Framework for Fair Regression}



\icmlsetsymbol{equal}{*}

\begin{icmlauthorlist}
\icmlauthor{Jack Fitzsimons}{ox}
\icmlauthor{AbdulRahman Al Ali}{nh}
\icmlauthor{Michael Osborne}{ox}
\icmlauthor{Stephen Roberts}{ox}
\end{icmlauthorlist}

\icmlaffiliation{ox}{University of Oxford, United Kingdom}
\icmlaffiliation{nh}{Northampton University, United Kingdom}

\icmlcorrespondingauthor{Jack Fitzsimons}{jack@robots.ox.ac.uk}

\icmlkeywords{Machine Learning, ICML, Algorithmic Fairness, Kernel Methods, Constrained Learning}

\vskip 0.3in
]



\printAffiliationsAndNotice{}  

\begin{abstract}
Fairness, through its many forms and definitions, has become an important issue facing the machine learning community. In this work, we consider how to incorporate group fairness constraints in kernel regression methods, applicable to Gaussian processes, support vector machines, neural network regression and decision tree regression. Further, we focus on examining the effect of incorporating these constraints in decision tree regression, with direct applications to random forests and boosted trees amongst other widespread popular inference techniques. We show that the order of complexity of memory and computation is preserved for such models and tightly bound the expected perturbations to the model in terms of the number of leaves of the trees. Importantly, the approach works on trained models and hence can be easily applied to models in current use and group labels are only required on training data.
\end{abstract}

\section{Introduction}
\label{intro}

As the proliferation of machine learning and algorithmic decision making continues to grow throughout industry, the net societal impact of them has been studied with more scrutiny. In the USA under the Obama administration, a report on big data collection and analysis found that ``big data technologies can cause societal harms beyond damages to privacy" \cite{united2014big}. The report feared that algorithmic decisions informed by big data may have harmful biases, further discriminating against disadvantaged groups. This along with other similar findings has led to a surge in research around algorithmic fairness and the removal of bias from big data.

The term \emph{fairness}, with respect to some sensitive feature or set of features, has a range of potential definitions. In this work, impact parity is considered. In particular, this work is concerned with group fairness under the following definitions as taken from \cite{gajane2017formalizing},

\textbf{Group Fairness:} \emph{A predictor $\mathcal{H}: X \rightarrow Y$ achieves fairness with bias $\epsilon$ with respect to groups $A, B \subseteq X$ and $O \subseteq Y$ being any subset of outcomes iff,}
    \[|\textnormal{P}\{\mathcal{H}(x_i) \in O | x_i \in A\} - \textnormal{P}\{\mathcal{H}(x_j) \in O | x_j \in B\}| \leq \epsilon\]

The above definition can also be described as statistical or demographical parity. Group fairness has found widespread application in India and the USA, where affirmative action has been used to address discrimination against caste, race and gender \cite{weisskopf2004affirmative, dumont1980homo, deshpande2017affirmative}. 

The above definition does not, unfortunately, have natural application to regression problems. One approach to get around this would be to alter the definition to bound the absolute difference between the respective marginal distributions over the output space. However, this is a strong requirement and may hinder the model's ability to model the function space appropriately. Rather, a weaker and potentially more desirable constraint would be to force the expectation of the marginal distributions over the output space to equate. Therefore,  statements such as \emph{``the average expected outcome for population $A$ and $B$ is equal"} would be valid.

The second issue encountered is that the generative distribution of groups $A$ and $B$ are generally unknown. In this work, it is assumed that the empirical distribution $p_A(x)$ and $p_B(x)$, as observed from the training set, is equal to or negligibly perturbed from the true generative distributions.

Combining these two caveats we arrive at the following definition,

\textbf{Group Fairness in Expectation:} \emph{A regressor $f(\cdot): X \rightarrow Y$ achieves fairness with respect to groups $A, B \subseteq X$ iff,}
    \[ \mathbb{E}[f(x_i) | x_i \in A] - \mathbb{E}[f(x_i) | x_j \in B] = 0 \]
    \[ \int \left(p_A(x) - p_B(x) \right) f(x) dx =  0 \]
    
There are many machine learning techniques with which \emph{Group Fairness in Expectation} constraints (GFE constraints) may be incorporated. While constraining kernel regression is introduced in Section 3, the main focus of the paper is examining decision tree regression and respective ensemble methods which build on decision tree regression such as random forests, extra trees and boosted trees due to their widespread use in industry and hence their extensive impact on society \cite{wu2008top}. The reason for this is to show that such an approach will not affect the order of computational or memory complexity of the model. 

The main contributions of this paper are,

\begin{itemize}
\item[I] The use of quadrature approaches to enforce GFE constraints on kernel regression with applications to Gaussian processes, support vector machines, neural network regression and decision tree regression, as outlined in Section 3.
\item[II] Incorporating these constraints on decision tree regression without affecting the computational or memory requirements, outlined in Sections 5 and 6.
\item[III] Deriving a tight bound for the variance of the perturbations due to the incorporation of GFE constraints on decision tree regression in terms of the number of leaves of the tree, outlined in Section 7.
\item[IV] Showing that these \emph{fair} trees can be combined into random forests, boosted trees and other ensemble approaches while maintaining fairness, as shown in Section 8.
\end{itemize}

\section{Related Work}

There have been primarily three branches of research towards the development of fair algorithms. The first is the data alteration approach, which endeavours to modify the original dataset in order to prevent discrimination or bias due to the protected variable \cite{luong2011k, kamiran2009classifying}. The second is an attempt to regularize such that the model is penalized for bias \cite{kamishima2011fairness, berk2017convex, calders2013controlling, calders2010three, raff2017fair}. Finally, the third endeavours to use post-processing to recalibrate and mitigate against bias \cite{kamiran2012decision, pleiss2017fairness}

More specifically to decision trees, discrimination aware decision trees have been introduced \cite{kamiran2010discrimination} for classification. The offer dependancy aware tree construction and leaf relabelling approach. Later, fair forests \cite{raff2017fair} introduced a further tree induction algorithm to encourage fairness. They did this by introducing a new gain measure to encourage fairness. However, the issue with adding such regularisation is two-fold. Firstly, discouraging bias via a regularising term does not make any guarantee about the bias of the post trained model. Secondly, it is hard to make any theoretical guarantees about the underlying model or the effect the new regulariser has had on the model.

The approach offered in this work seeks to perform model inference in a constrained space, leveraging basic theory from kernel quadrature such that the predicted marginal distributions are guaranteed to have equal means. By utilizing quadrature methods, it is also possible to derive bounds for the expected absolute perturbation induced by constraining the space. This is shown explicitly in Section 7.

\section{Constrained Kernel Regression}


We will first show how one can create such linear constraints on kernel regression models.  This work builds on the earlier contributions of \cite{jidling2017linearly}, where the authors examined the incorporation of linear constraints on Gaussian processes (GPs). Gaussian processes are a Bayesian kernel method most popular for regression. For a detailed introduction to Gaussian processes, we refer the reader to \cite{rasmussen2004gaussian}. However, for the reader unfamiliar with GPs specifically, they may simply think of a high dimensional Gaussian distribution parameterized by a kernel $K(\cdot, \cdot)$, with zero mean and unit variance without loss of generality. Given a set of inputs and respective outputs, $\{\emph{x}_i, \emph{y}_i\}_{i=1}^N$, split into training and testing sets,$\{\textnormal{x}_i, \textnormal{y}_i\}_{i=1}^n$ and $\{\bar{\textnormal{x}}_i, \bar{\textnormal{y}}_i\}_{i=1}^{N-n}$, inference is performed as,

\vspace{-4mm}

\[
\mathbb{E}[\bar{\textnormal{y}}] = K_{\bar{x}, x} K_{x, x}^{-1} \textnormal{y}
\]

\vspace{-5mm}

\[
\mathbb{V}[\bar{\textnormal{y}}] = K_{\bar{x}, \bar{x}} - K_{\bar{x}, x} K_{x, x}^{-1} K_{x,\bar{x}}
\]

\noindent where $K_{x, x}$ denotes the kernel matrix between training examples, $K_{\bar{x}, x}$ is the kernel matrix between the test and training examples and $K_{\bar{x}, \bar{x}}$ is the prior variance on the prediction point defined by the kernel matrix. Gaussian processes differ from high dimensional Gaussian distributions as they can model the relationships between points in continuous space, via the kernel function, as opposed to being limited to a finite dimension.

An important note is that any combination of Gaussian distributions via addition and subtraction is a closed space, that is to say, the sum of Gaussians is also Gaussian and so on. While this may at first appear trivial, it is, in fact, a very useful artefact. For example, let us assume there are two variables, $a$ and $b$, drawn from Gaussian distributions with mean and variance $\mu_a, \mu_b, \sigma_a^2, \sigma_b^2$ respectively. Further, assume that the correlation coefficient $\rho$ describes the interaction between the two variables. Then a new variable $c$, which is equal to the difference $a$ and $b$, is drawn from a Gaussian distribution with mean and variance,

\vspace{-4mm}

\[
\mu_c = \mu_a - \mu_b
\]

\vspace{-5mm}

\[
\sigma_c^2 = \sigma_a^2 + \sigma_b^2 - 2\rho\sigma_a\sigma_b
\]

\noindent We can thus write all three variables in terms of a single mean vector and covariance matrix,

\[
\mu =\begin{bmatrix}
     \mu_a - \mu_b\\
      \mu_a\\
     \mu_b\\
\end{bmatrix}
\]
\[
K = \begin{bmatrix}
    \sigma_a^2 + \sigma_b^2 - 2\rho\sigma_a\sigma_b & \sigma_a^2 - \rho\sigma_a\sigma_b & \sigma_b^2 - \rho\sigma_a\sigma_b\\
     \sigma_a^2 - \rho\sigma_a\sigma_b & \sigma_a^2 & \rho\sigma_a\sigma_b\\
     \sigma_b^2 - \rho\sigma_a\sigma_b& \rho\sigma_a\sigma_b& \sigma_b^2\\
\end{bmatrix}
\]

\noindent Given any two of the above observations, the third can be inferred exactly. We refer to this as a degenerate distribution as $K$ will naturally be low rank. If we observe that $\mu_a - \mu_b$ is equal to zero, we are thus constraining the distribution of $a$ and $b$. This can easily be extended to the relationship between sums and differences of more variables.

Bayesian quadrature \cite{o1991bayes} is a technique used to incorporate integral observations into the Gaussian process framework. Essentially, quadrature can be derived through an infinite summation and the above relationship between these summations can be exploited \cite{osborne2012active}.  An example covariance structure thus looks akin to,

\vspace{-4mm}

\[
K = \begin{tiny}\begin{bmatrix}
    \int \int p(x) K(x, x') p(x') \textnormal{d}x \textnormal{d}x' & \int p(x) K(x, x_0)  \textnormal{d}x & \int p(x) K(x, x_1)  \textnormal{d}x\\
     \int p(x) K(x, x_0)  \textnormal{d}x & K(x_0, x_0) & K(x_1, x_0) \\
     \int p(x) K(x, x_1)  \textnormal{d}x & K(x_1, x_0) & K(x_1, x_1)\\
\end{bmatrix}
\end{tiny}
\]

\noindent where $p(x)$ is some probability distribution over the domain of $x$, on which the Gaussian process is defined and against which the quadrature is performed against.

Reiterating the the motivation of this work; given two generative distributions $p_A(x)$ and $p_B(x)$ which subpopulations $A$ and $B$ of the data are generated from, we wish to constrain the inferred function $f(\cdot)$ such that,

\[
\int p_A(x) f(x) \textnormal{d}x = \int p_B(x) f(x) \textnormal{d}x.
\]

\noindent This constraint can be rewritten as,

\[
\int \left( p_A(x) - p_B(x) \right) f(x) \textnormal{d}x = 0,
\]

\noindent which allows us to incorporate the constraint on $f(\cdot)$ as an observation in the above Gaussian process. Let $q_{A,B}(x) = p_A(x) - p_B(x)$ be the difference between the generative probability distributions of $A$ and $B$, then by setting the corresponding observation as zero, the covariance matrix becomes,

\vspace{-4mm}

\[
\mkern-20muK = \begin{tiny}\begin{bmatrix}
    \int \int q_{A,B}(x) K(x, x') q_{A,B}(x') \textnormal{d}x \textnormal{d}x' & \int q_{A,B}(x) K(x, x_0) \textnormal{d}x & \int q_{A,B}(x) K(x, x_1)  \textnormal{d}x\\
     \int q_{A,B}(x) K(x, x_0)  \textnormal{d}x & K(x_0, x_0) & K(x_1, x_0) \\
     \int q_{A,B}(x) K(x, x_1)  \textnormal{d}x & K(x_1, x_0) & K(x_1, x_1)\\
\end{bmatrix}
\end{tiny}
\] 

We will refer to these as equality constrained Gaussian processes. Let us now turn to incorporate these concepts into decision tree regression.

\section{Trees As Kernel Regression}

Decision tree regression (DTR) and related approaches offer a white box approach for practitioners who wish to use them. These methods are among the most popular methods in machine learning \cite{wu2008top} in practice as they are generally intuitive even for those not from statistics, mathematics or computer science background. It is their proliferation, especially in businesses without machine learning researchers, that makes them of particular interest. 

DTR regress data by sorting them down binary trees based partitions in the input domain. The trees are created by recursively partitioning the domain of input along axis aligned splits determined by a given metric of the data in each partition, such as information gain or variance reduction. In this work, we will not consider the many possible techniques for learning decision trees, but rather assume that the practitioner has a trained decision tree model. For a more complete description of decision trees, the authors refer the readers to \cite{rokach2008data}.

For the purposes of this work, DTR can be described as a partitioning of space such that predictions are made by averaging the observations in the local partition, referred to as the leaves of the tree. As such, DTR has a very natural formulation as a degenerate kernel whereby,

\[
K(x_i, x_j) = \begin{cases}
      1, & \text{if}\ L(x_i) = L(x_j) \\
      0, & \text{otherwise}    
 \end{cases}
\]

\noindent where $L(\cdot)$ is the index of the leaf in which the argument belongs. The kernel hence becomes naturally block diagonal and the classifier / regressor written as,

\[
f(\bar{\textnormal{x}}) = \mathbb{E}[\bar{\textnormal{y}}] = K_{\bar{x}, x} K_{x, x}^{-1} \textnormal{y}
\]

\noindent with $K_{\bar{x}, x}$ denoting the vector of kernel values between $\bar{\textnormal{x}}$ and the observations, $K_{x, x}$ denotes the covariance matrix of the observations as defined by the implicit decision tree kernel and $\textnormal{y}$ denoting the values of the observations. 

It is worth also noting how one can also write the decision tree as a two-stage model: first by averaging the observations of associated with each leaf and then by using a diagonal kernel matrix to perform inference. Trivially, the diagonal kernel matrix acts only as a lookup and outputs the leaf average that corresponds to the point being predicted. Let us refer to this compressed kernel matrix approach as the \emph{compressed kernel representation} and the block diagonal variant as the \emph{explicit kernel representation}. 

\section{Fairness Constrained Decision Trees}

Borrowing concepts from the previous section on equality constrained Gaussian processes using Bayesian quadrature, decision trees may be constrained in a similar fashion. The first consideration to note is that we wish the constraint observation to act as a hard equality, that is to say noiseless. In contrast, we are willing for the observations to be perturbed in order to satisfy this hard equality constraint. To achieve this, let us add a constant noise term, $\sigma_{noise}^2$, to the diagonals of the decision tree kernel matrix. Similar to ordinary least squares regression, the regressor will now minimize the L2-norm of the error induced on the observations, conditioned on the equality constraint which is noise free. In the explicit kernel representation, this implies the minimum induced noise \emph{per observation}, whereas in compressed kernel representation this implies the minimum induced noise \emph{per leaf}. 

An important note is that the constraint is applied to the kernel regressor equations, hence the method is exact for regression trees or when the practitioner is concerned with relative outcomes of various predictions. However, in the case that the observations range between $[0,1]$, as is the case in classification, then we must renormalize the output to $[0,1]$. This no longer guarantees a minimum L2-norm perturbation and while potentially still useful, is not the focus of this work.

The second consideration is how to determine the generative probability distributions $p_A(x)$ and $p_B(x)$. Given the frequentist nature of decision trees, it makes sense to consider $p_A(x)$ and $p_B(x)$ as the empirical distributions of subpopulations $A$ and $B$ as described in Section 1. Thus the integral of the empirical distribution on a given leaf, $\int_{L_i} p_A(x) \textnormal{d}x$, is defined as the proportion of population $A$ observed in the partition associated with leaf $L_i$. We emphasise that how $p_A(x)$ and $p_B(x)$ are determined is not the core focus of this work and many approaches have merit. For example, a Gaussian mixture model could be used to model the input distribution in which case $\int_{L_i} p_A(x) \textnormal{d}x$ would equal the cumulative distribution of the generative PDF over the bounds defined by the leaf. This is demonstrated in the experimental section. Many other such models would also be valid and determining which method to use to model the generative distribution is left to the practitioner with domain expertise.

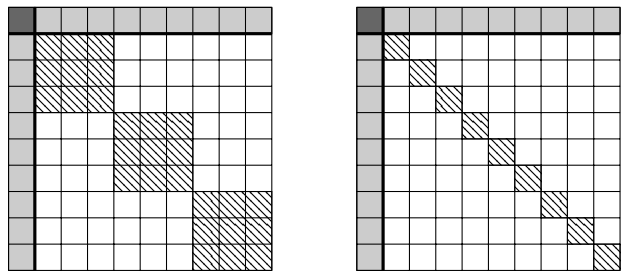
\begin{figure}
\label{fig:kernel}
\begin{center}
\begin{tikzpicture}[scale=.35]
  \begin{scope}
    \filldraw[fill=black!60!white] (0,9.) rectangle (1,10);
    
    \filldraw[fill=black!20!white] (1,9.) rectangle (10,10);
    \filldraw[fill=black!20!white] (0,0.) rectangle (1,9);
    

    \draw[pattern=north west lines, pattern color=black] (1,9.) rectangle (4,6.);
    \draw[pattern=north west lines, pattern color=black] (4,6.) rectangle (7,3);
    \draw[pattern=north west lines, pattern color=black] (7,3) rectangle (10,0);

    \draw (0, 0) grid (10, 10);
    \draw[very thick, scale=3] (3.333, 3) -- (0, 3);
    \draw[very thick, scale=3] (0.333, 0) -- (0.333, 3.333);
\end{scope}
\end{tikzpicture}\hspace{10mm}
\begin{tikzpicture}[scale=.35]
  \begin{scope}
    \filldraw[fill=black!60!white] (0,9.) rectangle (1,10);
    
    \filldraw[fill=black!20!white] (1,9.) rectangle (10,10);
    \filldraw[fill=black!20!white] (0,0.) rectangle (1,9);
    

    \draw[pattern=north west lines, pattern color=black] (1,9.) rectangle (2,8.);
    \draw[pattern=north west lines, pattern color=black] (2,8.) rectangle (3,7);
    \draw[pattern=north west lines, pattern color=black] (3,7) rectangle (4,6);
    \draw[pattern=north west lines, pattern color=black] (4,6.) rectangle (5,5.);
    \draw[pattern=north west lines, pattern color=black] (5,5.) rectangle (6,4);
    \draw[pattern=north west lines, pattern color=black] (6,4) rectangle (7,3);
    \draw[pattern=north west lines, pattern color=black] (7,3.) rectangle (8,2.);
    \draw[pattern=north west lines, pattern color=black] (8,2.) rectangle (9,1);
    \draw[pattern=north west lines, pattern color=black] (9,1) rectangle (10,0);

    \draw (0, 0) grid (10, 10);
    \draw[very thick, scale=3] (3.333, 3) -- (0, 3);
    \draw[very thick, scale=3] (0.333, 0) -- (0.333, 3.333);
\end{scope}
\end{tikzpicture}

\end{center}
\caption{Above is a visualization of a decision tree kernel matrix with marginal constraint, left in explicit representation and right in compressed representation. The dark cell in the upper left of the matrix is the double integrated kernel function with respect to the difference of input distributions which constrain the process. The solid grey row and column are single integrals of the kernel function. White cells have zero values and the dashed (block) diagonals are the kernel matrix between observations or leaves of the tree. We can note that the above, compressed representation kernel matrix is an arrowhead matrix, which we will exploit to create an efficient algorithm.}
\end{figure}


\section{Efficient Algorithm For Equality Constrained Decision Trees}

At this point, an equality constrained variant of a decision tree has been described, both in explicit representation and compressed representation. In this section, we will show that equality constraints on a decision tree do not change the computational or memory order of complexity. The motivation for considering the order of complexities is that decision trees are one of the more scalable machine learning models, whereas kernel methods such as Gaussian processes naively scale at $\mathcal{O}(n^3)$ in computation and $\mathcal{O}(n^2)$ in memory, where $n$ is the number of observations. While the approach presented in this work utilizes concepts from Bayesian quadrature and linearly constrained Gaussian processes, the model's usefulness would be drastically hindered if it no longer maintained the performance characteristics of the classic decision tree, namely computational cost, and memory requirements.

\subsection{Efficiently Constrained Decision Trees in Compressed Kernel Representation}

As Figure 1 shows, the compressed kernel representation of the constrained decision tree creates an arrowhead matrix. It is well known that the inverse of an arrowhead matrix is a diagonal matrix with a rank-1 update. Letting $D$ represent the diagonal principal sub-matrix with diagonal elements equal to one, $z$ being vector such that the $i^{th}$ element is equal to the relative difference in generative populations distributions for leaf $i$, $z_i = \int_{L_i} (p_A(x) - p_B(x)) dx$, then the arrowhead inversion properties state that,

\[
\begin{bmatrix}
    D & z\\
    z^T & 0\\
\end{bmatrix}^{-1}
=
\begin{bmatrix}
    D^{-1} & \textbf{0}\\
    \textbf{0}^T & 0\\
\end{bmatrix} +
\rho uu^T,
\]

\noindent with $\rho = - \frac{1}{z^TD^-1z}$ and $u = \begin{bmatrix}
    D^{-1}z\\
    -1\\
\end{bmatrix}$. Note that the integral of the difference between the two generative distributions when evaluated over the entire domain is equal to zero, as both $p_A(x)$ and $p_B(x)$ must sum to one by definition and hence their differences to zero. Returning to the equation of interest, namely $f(\bar{\textnormal{x}}) = K_{\bar{x}, x} K_{x, x}^{-1} \textnormal{y}$ with $\textnormal{y}$ as the average value of each leaf of the tree, and subbing in $K_{\bar{x}, x}$ as a vector of zeros with a one indexing the $j^{th}$ leaf in which the predicted point belongs to and is equal to zero, as it does not contribute to the empirical distributions, we arrive at,

\[
f(\bar{\textnormal{x}}) = \frac{1}{1 + \sigma_n^2} \left( \textnormal{y}_j + z_j \frac{\sum_i z_i y_i}{\sum_i z_i^2 } \right). 
\]

\noindent The term $\frac{1}{1 + \sigma_n^2}$ is the effect of the prior under the Gaussian process perspective, however, by post-multiplying by $(1 + \sigma_n^2)$, this prior effect can be removed. While relatively simple to derive, the above equation shows that only an additive update to the predictions is required to ensure group fairness in decision trees. Further, if the same relative population is observed for group $A$ and group $B$ on a single leaf $j$, then $z_j = 0$ and no change is applied to the original inferred prediction before the constraint is applied other than the effect of the noise. In fact, the perturbation to a leaf's expectation grows linearly with the bias in the population of the leaf.

From an efficiency standpoint, only the difference in generative distributions, $z$, needs to be stored which is an additional $\mathcal{O}(L)$ extra memory requirement and the update per leaf can be pre-computed in $\mathcal{O}(L)$. These additional memory and computational requirements are negligible compared to $\mathcal{O}(N)$ cost of the decision tree itself.


\subsection{Efficiently Constrained Decision Trees in Explicit Kernel Representation}
\label{blockDTupate}

Let us now turn our attention to the explicit kernel representation case, that is where the $D$ in the previous subsection is replaced with the block diagonal matrix equivalent. First let us state the \emph{bordering method}, a special case of the block diagonal inversion lemma,

\vspace{-5mm}

\[
\begin{bmatrix}
    D & z\\
    z^T & 0\\
\end{bmatrix}^{-1}
=
\begin{bmatrix}
    D^{-1} - \rho D^{-1}z z^TD^{-1}  & \rho D^{-1} z\\
    \rho z^TD^{-1} & - \rho\\
\end{bmatrix}.
\]

\noindent with $\rho = - \frac{1}{z^TD^-1z}$ once again. Substituting this into the kernel regression equation once more we find,

\vspace{-5mm}

\begin{equation*}
\begin{split}
f(\bar{\textnormal{x}}) &= \begin{bmatrix}
    \mathbb{I}_j\\
    0\\
\end{bmatrix}\begin{bmatrix}
    D & z\\
    z^T & 0\\
\end{bmatrix}^{-1}\begin{bmatrix}
    \textnormal{y}\\
    0\\
\end{bmatrix}\\
&=
\begin{bmatrix}
    \mathbb{I}_j\\
    0\\
\end{bmatrix}\begin{bmatrix}
    D^{-1} - \rho D^{-1}z z^TD^{-1}  & \rho D^{-1} z\\
    \rho z^TD^{-1} & - \rho\\
\end{bmatrix}\begin{bmatrix}
    \textnormal{y}\\
    0\\
\end{bmatrix}
\end{split}
\end{equation*}

\noindent where $\mathbb{I}_j$ denotes a vector of zeros with ones placed in all elements relating to observations in the same leaf. Expanding the above linear algebra,


\[f(\bar{\textnormal{x}}) = \mathbb{I}_jD^{-1}\textnormal{y} + \rho \mathbb{I}_jD^{-1} z z^TD^{-1}\textnormal{y}.
\]

\noindent As D is a block-diagonal matrix, it is straight forward to show $\rho = - \left( \sum_{j \in L} \frac{m_j z_j^2}{m_j + \sigma_n^2} \right)^{-1}$ where $j$ is iterating over the set of leaves. Note that when $m_j = 1$ for all $j$ we arrive at the same value for $\rho$ as we did in the previous subsection. We can continue to apply this result to the other terms of interest,

\vspace{-6mm}

\begin{equation*}
\begin{split}
X_1 &= \mathbb{I}_jD^{-1}\textnormal{y} = \frac{m_j \textnormal{y}_j}{m_j + \sigma_n^2}\\
X_2 &= \mathbb{I}_jD^{-1} z = \frac{m_j z_j}{m_j + \sigma_j^2}\\
X_3 &= z^TD^{-1}\textnormal{y} = \sum_{j \in L}  \frac{m_j z_j \textnormal{y}_j}{m_j + \sigma_n^2} \\
\end{split}
\end{equation*}

\noindent where $y_j$ is once again the average output observation over leaf $j$. The terms have been labelled $X_1$, $X_2$ and $X_3$ for shorthand. The computation time for the three terms, along with $\rho$, can be computed in linear time with respect to the size of the data, $\mathcal{O}(n)$, and can be pre-computed ahead of time, hence not affect the computational complexity of a standard decision tree. Once again only $z_j$ and $m_j$ have to be stored for each leaf and hence the additional memory cost is only $\mathcal{O}(L)$. As such we can simplify the full expression for the expected outcome as,

\[f(\bar{\textnormal{x}}) = X_1 + \rho X_2 X_3.
\]

\section{Expected Perturbation Bounds}

In imposing equality constraints on the models the inferred outputs become perturbed. In this section, the expected magnitude of the perturbation is analyzed for the compressed kernel representation. We define the perturbation due to the equality constraint, not due to the incorporation of the noise, as,

\[
\epsilon = z_j  \frac{\sum_i z_i y_i}{\sum_i z_i^2}.
\] 

\noindent \textbf{Theorem 1} \emph{Given a decision tree with $L$ leaves, with expected value of leaf observations denoted by the vector $y \in \mathbb{R}^{L}$ normalized to have zero mean and unit variance and leaf frequency imbalance denoted as  $z \in \mathbb{R}^{L}$ the expected variance induced by the perturbation due to the incorporating a Group Fairness in Expectation constraint is bounded by,}

\[\mathbb{E}[\epsilon^2] \leq \frac{1}{L}\]

\textbf{Proof 1} As the expectation of $z_j$ is zero due to it being the difference of two probability distributions, the variance is equal to the expectation of $\epsilon^2$,

\[
        \mathbb{E}[\epsilon^2] = \frac{1}{L} \sum_{j=1}^L z_j^2 \frac{(\sum_i z_i y_i)^2}{(\sum_i z_i^2)^2}
         = \frac{1}{L} \|z\|_1^2  \frac{\|z\|_2^2 (\bar{z}^Ty)^2}{\|z\|_2^4}
\]

\noindent with $\bar{z}$ equal to $z$ after normalization. By Lemma 1, the expectation of the dot product $(\bar{z}^Ty)^2$ is equal to $\frac{1}{L}$. Further, the 2-norm of $z$ can be cancelled from the numerator and denominator. Finally, using the $L_1,L_2$ norm inequality, $\|z\|_2 \leq \|z\|_1 \leq \sqrt{L}\|z\|_2$, we can then tightly bound the worst case introduced variance as,

\[
\mathbb{E}[\epsilon^2] = \frac{1}{L^2} \|z\|_1^2 \|z\|_2^{-2} \leq \frac{1}{L}
\]

\textbf{Lemma 1} \emph{Given two vectors $y, \bar{z}$ uniformly distributed on the unit hypersphere $\mathcal{S}^{L-1}$, the expectation of their dot product is zero and variance,}

\[
\mathbb{E}[(\bar{z}^T y)^2] = \frac{1}{L}.
\]

\textbf{Proof 2} As the inner product is rotation invariant when applied to both $\bar{z}$ and $y$, let us denote the vector $\bar{z}$ as $[1, 0, \dots, 0]$ without loss of generality. The first element of the vector $y$, denoted by $y_0$, will thus be equal to $\bar{z}^Ty$. The probability density mass of the random variable $y_0$ is proportional to the surface area lying at a height between $y_0$ and $y_0 + dy_0$ on the unit hypersphere. That proportion occurs within a belt of height $dy_0$ and radius $\sqrt{1 - y_0^2}$, which is a conical frustum constructed out of an $S^{L-2}$ of radius $\sqrt{1 - y_0^2}$, of height $dy_0$, and slope $\frac{1}{\sqrt{1 - y_0^2}}$. Hence the probability is proportional to,

\[
P(y_0) \sim \frac{(\sqrt{1 - y_0^2})^{D-2}}{\sqrt{1 - y_0^2}}dy_0 = (1 - y_0^2)^{\frac{L-3}{2}}
\]

\noindent Substituting $u = \frac{y_0 + 1}{2}$ we find that,

\begin{equation*}
    \begin{split}
        P(u)du &\sim (1 - (2u -1)^2)^{\frac{L-3}{2}} d(2u - 1)\\
        &= 2^{L-2} (u - u^2)^{\frac{L-3}{2}} du\\
        &= 2^{L-2} u^{\frac{L-1}{2} - 1} (u - 1)^{\frac{L-1}{2} - 1} du.\\
    \end{split}
\end{equation*}

\noindent Note that this last simplification of $P(u)$ is equal to the probability density function of the Beta distribution with both shape parameters equal $\alpha = \beta = \frac{L-1}{2}$. The variance of the Beta distribution is,

\[\frac{\alpha \beta}{(\alpha + \beta)^2 (\alpha + \beta + 1)} = \frac{1}{4L}.\]

\noindent Rescaling to find the variance of $y_0$ we arrive at $\frac{1}{L}$. As the expectation of $\mathbb{E}[\bar{z}^T y] = 0$ due to the properties of symmetry, $\mathbb{E}[(\bar{z}^T y)^2] = \frac{1}{L}$.

This is an interesting result as it implies that if the model is not exploiting biases in the generative distribution evenly across all of the leaves of the tree, that is to say, $\|z\|_1 = \sqrt{L} \|z\|_2$, then the resulting predictions will receive the greatest expected absolute perturbation when averaged over all possible $y$.

For the explicit kernel representation, the expected absolute perturbation bound can be analysed whereby each leaf holds an even number of observations. In such a scenario $m_i = m$ is equal for all leaves $i \in 1,\dots,L$. Substituting this into the equations for $\rho, X_2$ and $X_3$ we can find that the bounded expected perturbation is equal to,

\vspace{-2mm}

\[
\mathbb{E}[\epsilon^2] \leq \left( \frac{\sigma_{n}^2}{m + \sigma_{n}^2} \right)^2 \frac{1}{L}
\]

\vspace{-2mm}

\noindent For the sake of conciseness the full derivation of the above is left to the reader but follows the same steps as the compressed kernel representation.

\section{Combinations Of Fair Trees}

While it is intuitive to say that ensembles of trees with GFE constraints preserve the GFE constraint, however, for the sake of completeness this is now shown more formally. Random forests \cite{breiman2001random}, extremely random trees (ExtraTrees) \cite{geurts2006extremely} and tree bagging models \cite{breiman1996bagging} combine tree models by averaging over their predictions. Denoting the predictions of the trees at point $x$ as $f_i(x)$ for each $i \in 1,\dots, T$, where $T$ is the number of trees, we can easily show that the combined difference in expectation marginalised over the space is equal to zero,

\vspace{-5mm}

\begin{equation*}
\begin{split}
0 &=  \int (p_A(x) - p_B(x)) \sum_{i=1}^T f_i(x) \textnormal{d}x\\
 & = \sum_{i=1}^T \int (p_A(x) - p_B(x))  f_i(x) \textnormal{d}x     = \sum_{i=1}^T 0    \\
\end{split}    
\end{equation*}

It can also be easily shown that modelling residual errors of the trees with other fair trees, such as is the case for boosted tree models \cite{elith2008working}, results in fair predictors also. These concepts are not limited to tree methods either and the core concepts set out in this paper of constraining kernel matrices can have applications in models such as deep Gaussian process models \cite{damianou2013deep}.

\section{Experiments}

\subsection{Synthetic Demonstration}

The first experiment is a visual demonstration to better communicate the validity of the approach. The models examined are ExtraTrees, Gaussian processes and a single hidden layer perceptron. They endeavour to model an analytic function, $f(x) = x\cos(\alpha x^2) + \sin(\beta x)$, with observations drawn from two beta distributions, $p_A(x)$ and $p_B(x)$ respectively. The parameters of the two beta distribution are,

\begin{table}[h!]
\begin{center}
\begin{tabular}{c|c c c}
Group& $p_A(x)$ &$p_B(x)$\\
\hline
$\alpha$&2&3\\
$\beta$&3&2\\
\end{tabular}
\end{center}
\caption{Parameters of Beta distributions used to create synthetic samples.}
\end{table}

\begin{figure}[h!]
\centering
  \includegraphics[width=0.43\textwidth]{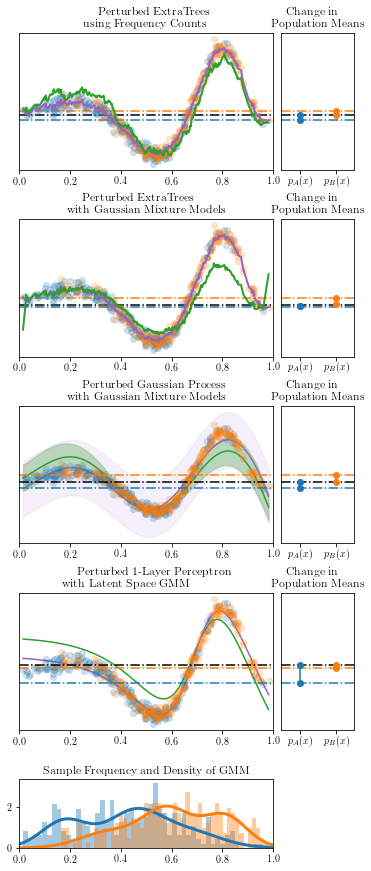}
  \caption{Above shows synthetic data of two populations, $p_A(x)$ (blue) and $p_B(x)$ (orange). The main plots show the observations and the perturbation to the respective models. Purple functions identify the original inferred functions and green indicates the fair perturbed inferred functions. Below the main plots show a normalised histogram of the observations for the $p_A(x)$ and $p_B(x)$ populations respectively along with the PDF of the Gaussian mixture model of their respective densities. To the right shows how the expected mean of the two populations have been perturbed to be equal.}
  \label{fig:demo}
\end{figure}

\noindent Figure~\ref{fig:demo} shows the effect of perturbing the models using the approach presented to constrain the expected means of the two populations. The figure shows the greater disparity between $p_A(x)$ and $p_B(x)$, the greater the perturbation in the inferred function. Both the compressed and explicit kernel representation lead to very similar plots for the tree-based models, so only the compressed kernel representation algorithm has been shown for conciseness. Note in the case of the ExtraTrees model, each tree was individually perturbed before being combined. Further, in the case of the perceptron, a GMM was fit to the data in the inferred latent space rather than in the original input space.

A downside to group fairness algorithms more generally, as pointed out in \cite{luong2011k}, is that candidate systems which impose group fairness can lead to qualified candidates being discriminated against. This can be visually verified as the perturbation pushes down the outcome of many orange points below the total population mean in order to satisfy the constraint. By choosing to incorporate group fairness constraints the practitioner should be aware of these tradeoffs. 

\subsection{ProPublica Dataset - Racial Biases}

Across the USA, judges, probation and parole officers are increasingly using algorithms to aid in their decision making. The ProPublica dataset \footnote{\url{https://www.propublica.org/datastore/dataset/compas-recidivism-risk-score-data-and-analysis}} contains data about criminal defendants from Florida in the United States. It is the Correctional Offender Management Profiling for Alternative Sanctions (COMPAS) algorithm \cite{dieterich2016compas} which is often used by judges to estimate the probability that a defendant will be a recidivist, a term used to describe re-offenders. However, the algorithm is said to be racially biased against African-Americans \cite{dressel2018accuracy}. In order to highlight the proposed algorithm, we first endeavour to use a random forest to approximate the decile scoring of the COMPAS algorithm and then perturb each tree to remove any racial bias from the system. 

The two subpopulations we consider constraining are thus African-American and non-African-American. We encode the COMPAS algorithms decile score into an integer between zero and ten such that minimizing $L_2$ perturbation is an appropriate objective function. The fact the decile scores are bounded in $[0,10]$ was not taken into account. The random forest used 20 decision trees as base estimators and the explicit kernel representation version of the algorithm was used for the sake of demonstrative purposes. 

\begin{figure}[h!]
\centering
  \includegraphics[width=0.4\textwidth]{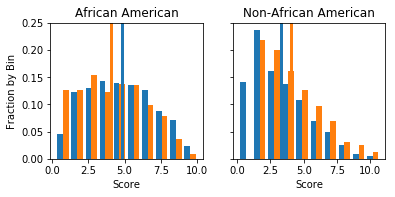}
  \caption{The above figure shows the output distribution of decile scores for African Americans and non-African Americans before (blue) and after (orange) the mean equality constraint was applied. We can see that the respective means (vertical lines) become approximately equal after the inclusion of the constraint using the empirical input distribution.}
  \label{fig:demo2}
\end{figure}

Figure 3 presents the marginal distribution of predictions on a 20\% held out test set before and after the GFE constraint was applied. It is visible that both the expected outcome for African Americans is decreased and for non-African Americans is increased. Notice that while the means are equal to the structure of the two of distributions are quite different, indicating that GFE constraints still allow greater flexibility than more strict group fairness such as that described in Section 1. The root square difference between the predicted points before and after perturbation was 0.8. Importantly, the GFE constraint described in this work was verified numerically with the average outputs recorded as,

\begin{table}[h!]
\begin{center}
\begin{center}
\begin{tabular}{l |c r}
&Unconstrained& Constrained\\
\hline
African Am.&4.82&4.41\\
non-African Am.&3.26&4.41\\
\end{tabular}
\end{center}
\end{center}
\caption{Mean score before and after GFE perturbation. }
\end{table}

\subsection{Intersectionality: Illinois State Employee Salaries}

The Illinois state employee salaries\footnote{\url{https://data.illinois.gov/datastore/dump/1a0cd05c-7d17-4e3d-938d-c2bfa2a4a0b1}} since 2011 can be seen to have a gender bias and bias between veterans and non-veterans. The motivation of this experiment was to show how we can deal with intersectionality issues (multiple compounding constraints) such as if one wished to predict a fair salary for future employees based on current staff. Gender labels were inferred using the employees' first names, parsed through the \emph{gender-geusser} python library. GFE constraints were applied between all intersections of gender and veteran / non-veterans, the marginals of gender and the marginals of veteran / non-veterans. Figure 4 visualizes the perturbations to the marginals of each demographic intersection due to the GFE constraints. The train-test split was set as 80\%-20\% and the incorporation of the GFE constraints increase the root mean squared error from \$12,086 to \$12,772, the cost of fairness. The only difference to allow for intersectionality is the $z$ is no longer a vector, but rather a matrix with a column for each constraint. Thus,

\[
f(\bar{\textnormal{x}}) = \textnormal{y}_j + z_j (z^T z)^{-1} z^T y. 
\]

\begin{figure}[h!]
\centering
  \includegraphics[width=0.45\textwidth]{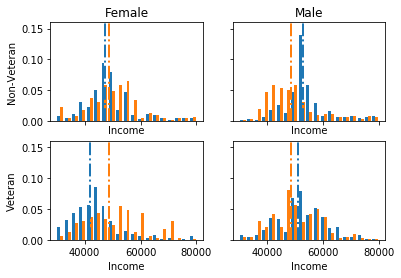}
  \caption{The above figure shows the effect of multiple GFE constraints acting on a single regression task. Blue signifies the original model and orange the perturbed model. The dashed horizontal lines signify the mean before and after perturbation.}
  \label{fig:demo3}
\end{figure}

\section{CONCLUSION}

This work offers an easily implementable approach to constrain the means of kernel regression which has direct applicability to random forest regression, boosted trees and other tree-based ensemble models.

\bibliographystyle{abbrvnat}
\bibliography{biblio}

\end{document}